\documentclass[letterpaper]{article}
\usepackage[submission]{aaai23}
\usepackage{times}
\usepackage{helvet}
\usepackage{courier}
\usepackage[hyphens]{url}
\usepackage{graphicx}
\usepackage{natbib}
\usepackage{caption}
\usepackage{algorithm}
\usepackage{algorithmic}

\usepackage{newfloat}
\usepackage{listings}
\DeclareCaptionStyle{ruled}{labelfont=normalfont,labelsep=colon,strut=off} % DO NOT CHANGE THIS
\floatstyle{ruled}
\newfloat{listing}{tb}{lst}{}
\floatname{listing}{Listing}

\usepackage{array}
\usepackage{booktabs}
\newcolumntype{P}[1]{>{\centering\arraybackslash}p{#1}}
\usepackage{multirow}
\usepackage{amssymb}
\usepackage{amsmath}

\title{One-Shot Video Inpainting}
\author{Sangjin Lee\equalcontrib \quad Suhwan Cho\equalcontrib \quad Sangyoun Lee
}

\affiliations{
    Yonsei University
}

\begin{document}
\maketitle

\begin{abstract}
Recently, removing objects from videos and filling in the erased regions using deep video inpainting (VI) algorithms has attracted considerable attention. Usually, a video sequence and object segmentation masks for all frames are required as the input for this task. However, in real-world applications, providing segmentation masks for all frames is quite difficult and inefficient. Therefore, we deal with VI in a one-shot manner, which only takes the initial frame's object mask as its input. Although we can achieve that using naive combinations of video object segmentation (VOS) and VI methods, they are sub-optimal and generally cause critical errors. To address that, we propose a unified pipeline for one-shot video inpainting (OSVI). By jointly learning mask prediction and video completion in an end-to-end manner, the results can be optimal for the entire task instead of each separate module. Additionally, unlike the two-stage methods that use the predicted masks as ground truth cues, our method is more reliable because the predicted masks can be used as the network’s internal guidance. On the synthesized datasets for OSVI, our proposed method outperforms all others both quantitatively and qualitatively.
\end{abstract}

\section{Introduction} 
\begin{figure}[t]
	\centering
	\includegraphics[width=1\linewidth]{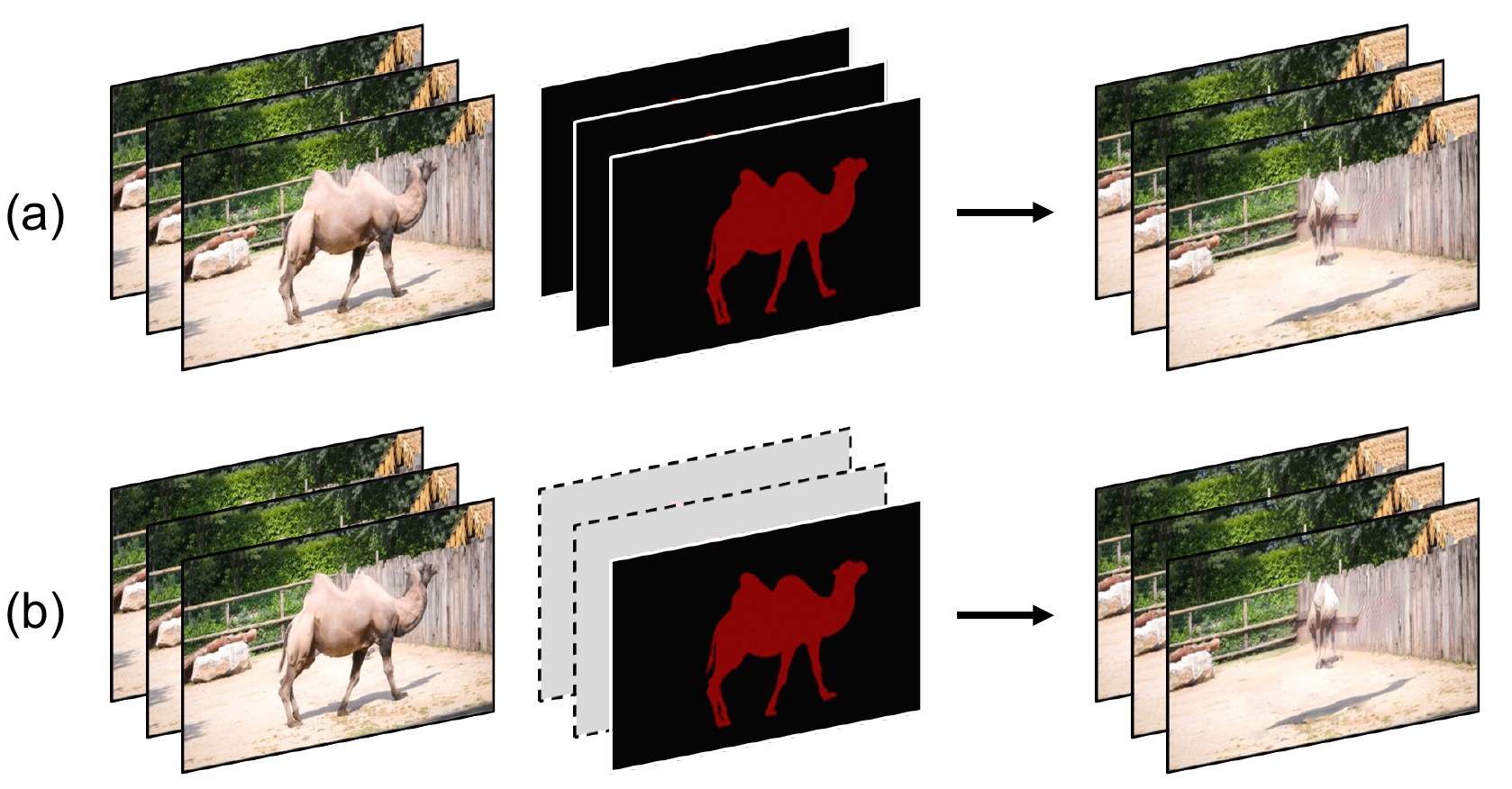}
    \caption{Visualized comparison between the definitions of (a) conventional video inpainting and (b) one-shot video inpainting.}
    \label{figure1}
\end{figure}

Video inpainting (VI) is a task that aims to remove a designated object in a given video sequence and fill in that area with plausible content. General approaches for VI take a video sequence and object segmentation masks for all frames as their input. However, in real-world applications, obtaining object segmentation masks for every frame is usually difficult and labor-intensive. To address this issue, we expand on an approach that is better suited to practical scenarios, namely one-shot video inpainting (OSVI). Compared to a conventional VI that requires full-frame segmentation masks as shown in Figure~\ref{figure1}~(a), OSVI requires only the initial frames's object mask and internally predicts those of subsequent frames while filling the missing content for all frames; this is depicted in Figure~\ref{figure1}~(b).

OSVI can be achieved by sequentially connecting a video object segmentation (VOS) network and a VI network as a two-stage pipeline. Using the initial frame object annotation, the VOS network first generates that object's entire frame masks. Then, using full-frame object masks as its input, the VI network fills in the object regions with plausible content that befits the background. However, this two-stage approach suffers from critical issues when directly applied to OSVI. Here, we will discuss this approach’s two main problems. First, since the entire network is not trained in an end-to-end manner, the results are optimal for each separate module but not as a whole. This property that two different networks are trained based on respective objectives leads to sub-optimal results for OSVI. Second, existing VI algorithms assume that the object masks received as input are always accurate. Therefore, if the predicted masks generated from a VOS model, which could be inaccurate, are provided as their input, the system will severely break down. This phenomenon will be particularly more drastic for flow-based VI methods where reference pixels are directly warped to the target region.

To address these concerns, we propose an end-to-end learnable baseline network that consists of mask prediction and video completion modules sharing a single encoder's embedded features. Unlike existing VI algorithms, we erase the object region at the feature level by adopting a novel transformer architecture. Our proposed method demonstrates its effectiveness for OSVI on synthetic datasets both quantitatively and qualitatively. Compared with two-stage methods, which are mainly naive combinations of existing VOS and VI methods, our method surpasses all of them by a large margin.

Our main contributions can be summarized as follows:
\begin{itemize}
\item We expand VI to OSVI, which refers to erasing a designated object in a video sequence only using a single frame annotation.
\item We propose a novel end-to-end learnable algorithm that can handle OSVI more effectively when compared to existing two-stage approaches.
\item Our proposed method outperforms all existing methods on synthetic datasets.
\end{itemize}

\section{Related Work}
\noindent\textbf{Object removal.} Due to its usefulness in diverse vision applications, such as video editing, object removal has been of interest recently. Shetty~\textit{et al.}~\cite{Auto2018adversarial} propose a GAN-based two-stage network composed of a mask generator and an image inpainter for object removal in an image. DynaFill~\cite{DynaFill} presents a new type of network using depth completion on the inpainted image. The depth map predicted from the inpainted image of a previous frame is used as guidance for current frame image inpainting. AutoRemover~\cite{autoremover} takes masks as its input and detects shadows using them. By using object masks and extracted shadow masks, the network can erase their shadows as well as objects in a video. VORNet~\cite{VORNet} fills the missing object regions only with box annotations as weak supervision to reduce annotation efforts.

\vspace{1mm}
\noindent\textbf{Video object segmentation.} Semi-supervised VOS is a pixel-level classification task that tracks and segments an arbitrary target object in a video. To efficiently handle an arbitrary target object, early works, such as OSVOS~\cite{OSVOS}, OnAVOS~\cite{OnAVOS} and OSVOS-S~\cite{OSVOS-S}, are based on online learning that trains a network during test time. However, since online learning is impractical for real-world applications because of its huge computational cost at test time, recent methods are based on feature matching that compares query frame features to reference frame features without online learning. VideoMatch~\cite{VideoMatch} extracts features from an initial frame and a query frame and then matches them at pixel-level using a soft matching layer. FEELVOS~\cite{FEELVOS} extends the use of pixel-level matching by employing initial and previous frames as reference ones; it also uses the matching output as the network's internal guidance. CFBI~\cite{CFBI} improves FEELVOS by employing background matching as well as foreground matching. To fully utilize the information from all past frames as well as initial and previous frames, STM~\cite{STM} proposes the space--time memory network, in which query frame features are densely matched to the memory features built from all past frames, covering all space--time locations. AFB-URR~\cite{AFB-URR} and XMem~\cite{XMem} improves the memory construction scheme to design a model robust to long videos.

\begin{figure*}[t]
    \centering
	\includegraphics[width=0.9\linewidth]{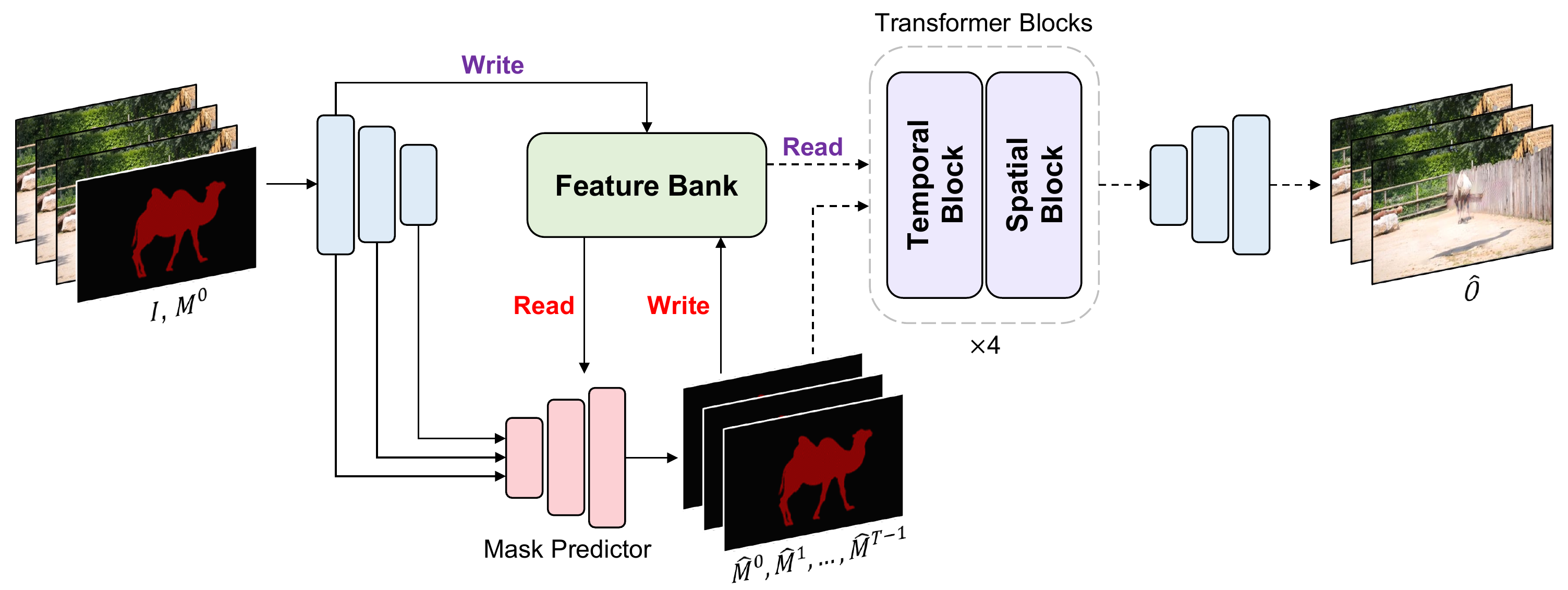}
    \caption{Overall architecture of our proposed algorithm. Solid lines indicate an operation is performed every frame, and dotted lines indicate an operation is performed after all solid line operations are done. Dotted lined operation takes a full-frame information as its input and generates full-frame predictions as its output at once. Red and purple letters refer to the information flows of the mask prediction and video completion, respectively.}
    \label{figure2}
\end{figure*}

\vspace{1mm}
\noindent\textbf{Video inpainting.} Early works~\cite{huang, space} use patch-based optimization to restore missing regions with valid regions for VI. However, these methods have some limitations. For example, they face difficulty in representing dynamic motions while maintaining temporal consistency, in addition to requiring high computational costs. To address these issues, various deep learning-based methods have been proposed within the past few years. These studies can be divided into two categories: patch-based approaches~\cite{CPNet, OPN, STTN, short, liu2021fuseformer} and flow-based approaches~\cite{VINet, DFVI, FGVC, TSAM, li2022towards}. Patch-based approaches aim to fill missing regions using direct pixel-level feature matching. CPNet~\cite{CPNet} uses an alignment network to fill missing regions by using the different object positions with background matching. OPN~\cite{OPN} proposes a similarity matching network that extracts and compares key and value features for each frame. It also suggests a post-processing network called the TCN, which uses a recurrent structure to increase temporal consistency. STTN~\cite{STTN} is composed of an encoder--decoder architecture and multi-layer multi-head spatio-temporal transformers trained using a 3D discriminator for temporal consistency. BSCA~\cite{short} proposes a boundary-aware short-term module for adjacent frame matching and a dynamic long-term context aggregation module for temporal consistency. FuseFormer~\cite{liu2021fuseformer} develops the transformer-based algorithms by proposing a method of efficiency patch splitting and composing. By contrast, flow-based approaches exploit the optical flow map for VI. VINet~\cite{VINet} proposes a flow warping-based recurrent network that refers to adjacent frames. DFVI~\cite{DFVI} restores flow maps extracted using a pre-trained flow estimation network by performing forward--backward pixel propagation. FGVC~\cite{FGVC} presents a flow--edge completion technique for the sharpness of flow map edges, a color propagation in the gradient domain for reducing warping error, and a non-local flow to resolve the chronic problem of flow maps; the latter namely being that they cannot obtain clearness when adjacent frames lack in information. TSAM~\cite{TSAM} proposes a method that applies flow warping to the feature level based on the DNN architecture. E2FGVI~\cite{li2022towards} shows the problems of previous flow-based pipelines and solves them by presenting the unified pipeline in an end-to-end manner.

\section{Proposed Approach}
\subsection{Problem Formulation}
Let us denote a video as $I:=\{I^i\in [0,255]^{3\times H0\times W0}~|~i=0, 1, ...T-1\}$, ground truth object masks as $M:=\{M^i\in \{0,1\}^{H0\times W0}~|~i=0, 1, ...T-1\}$, and a ground truth clean video as $O:=\{O^i\in [0,255]^{3\times H0\times W0}~|~i=0, 1, ...T-1\}$. Our goal is to generate the predicted object masks $\hat{M}$ and clean video $\hat{O}$, where $I$ and $M^0$ are provided as input.

\subsection{Network Overview}
An overview of our proposed network is shown in Figure~\ref{figure2}. It consists of two modules, a mask prediction module and a video completion module. First, the mask prediction module predicts the segmentation masks for all frames using the segmentation mask from the initial frame and the features extracted from each frame. The predicted masks and the extracted features are stored in a feature bank for reuse. As the semantic cues obtained from the mask prediction module are transferred to the video completion module without gradient disconnection, the entire network can be trained in an end-to-end manner.

\subsection{Mask Prediction Module}
\noindent\textbf{Memory construction.} To propagate the object information of the initial frame, i.e., ground truth segmentation mask, we employ a memory-based architecture as in STM~\cite{STM}. From $I^0$, we first extract base features $X^0$. Then, key features $K^0\in \mathbb{R}^{C_K\times HW}$ and value features $V^0 \in \mathbb{R}^{C_V\times HW}$ can be obtained as 
\begin{eqnarray}
    &K^0 = Conv(X^0)~,\\
    &V^0 = Conv(X^0 \oplus down(M^0))~,
\end{eqnarray}
where $\oplus$ and $down$ indicate channel concatenation and adaptive spatial pooling, respectively. The extracted key and value features are then stored in memory to be used for future frame prediction.

One of the memory architecture's main advantages is its ability to simultaneously store an arbitrary number of frames in memory. To fully exploit dense cues of a video, we store the key and value features of every five frames in the memory as well as the initial frame during the inference. Assuming $N$ frames are stored in the memory, the memory key and value features can be defined as $K_M\in \mathbb{R}^{C_K\times NHW}$ and $V_M\in \mathbb{R}^{C_V\times NHW}$, respectively.

\vspace{1mm}
\noindent\textbf{Memory read.} Based on the accumulated information (memory key and value features) obtained from past frames, the goal is to predict the query frame mask. To this end, we first extract query key features $K_Q\in \mathbb{R}^{C_K\times HW}$ from the query frame image and compare them to memory key features $K_M$ to calculate visual similarity. Then, we transfer the object cues, i.e., the memory value features $V_M$, to the query frame and the query frame value features $V_Q\in \mathbb{R}^{C_V\times HW}$ can be obtained. This process is identical to popular self-attention mechanism~\cite{attention}, and can be represented as
\begin{eqnarray}
    &sim = Softmax(K_M^T \otimes K_Q)~,\\
    &V_Q = V_M \otimes sim~,
\end{eqnarray}
where $Softmax$ and $\otimes$ indicate a query-wise softmax operation and matrix multiplication.

\vspace{1mm}
\noindent\textbf{Mask prediction.} As $V_Q$ contains object cues of the query frame, object segmentation masks $\hat{M}$ can be obtained by decoding it. To do that, we gradually increase the spatial size using the low-level features extracted from the encoder via skip connections. To obtain better feature representations, we add a CBAM~\cite{CBAM} layer after every skip connection.

\begin{figure*}[h]
\centering
	\includegraphics[width=0.9\linewidth]{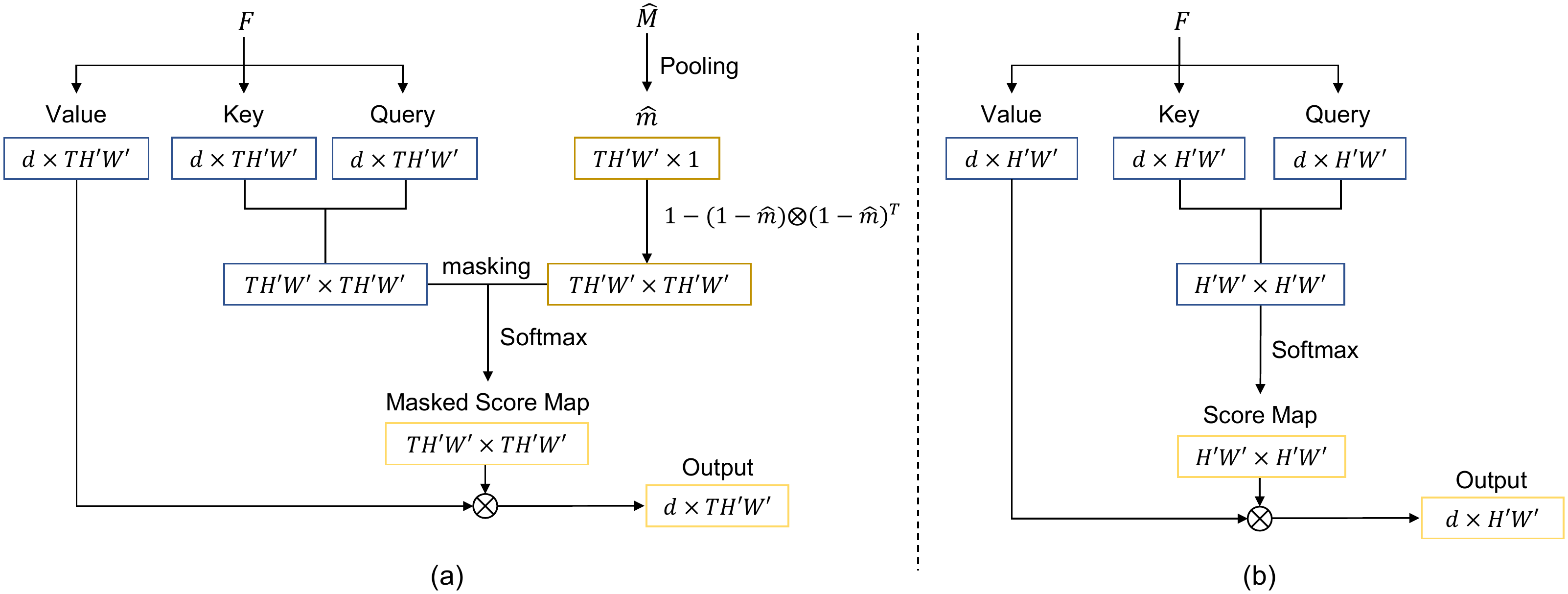}
    \caption{Visualized pipelines of (a) masked multi-head attention and (b) spatial multi-head attention.}
    \label{figure3}
\end{figure*}

\subsection{Video Completion Module}
\noindent\textbf{Masked multi-head attention.} In our framework, visual features are extracted from a video sequence and stored in the feature bank to be used for both mask prediction and video completion modules. Unlike conventional VI methods where object features are excluded, the features in our model contain object information since they should also be used to track a designated object. Considering object features may be referenced when filling the missing region, this will severely degrade the performance of the system. To address this issue, we propose a novel masked multi-head attention (MMHA) for transformer architecture.

\begin{figure}[t]
\centering
	\includegraphics[width=0.5\linewidth]{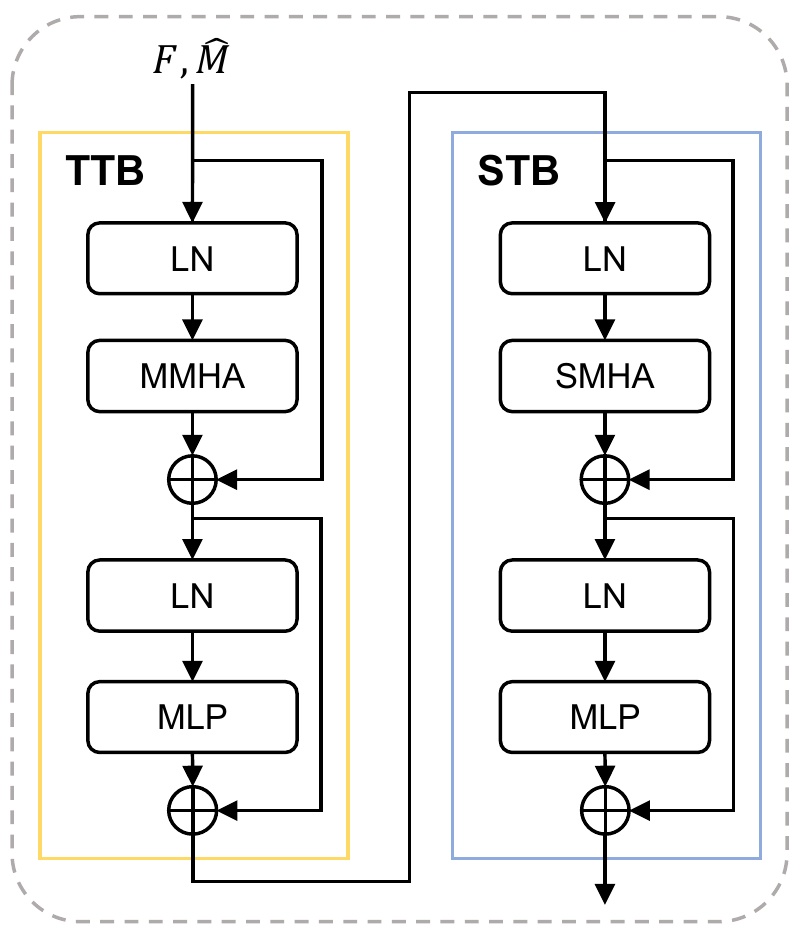}
    \caption{Architecture of our transformer block.}
    \label{figure4}
\end{figure}

The MMHA's visualized workflow is shown in Figure~\ref{figure3}~(a). Given that $F$ represents input features of a transformer block and $\hat{m}$ is a downsampled version of $\hat{M}$, we can obtain a mask guidance matrix $G \in [0,1]^{TH'W'\times TH'W'}$ as $1-(1-\hat{m}) \otimes (1-\hat{m})^T$, after reshaping $\hat{m}$ to $TH'W'\times 1$. Then, the mask guidance matrix is masked to a key--query similarity map to remove the object region from consideration. However, as a softmax operation comes after this process, we adopt a value substitution operation instead of a simple multiplication. This process can be represented as
\begin{equation}
masking(s, G) = 
\begin{cases}
-\infty &G < 0.5 \\ s & otherwise~,
\end{cases}
\end{equation}
where $s \in \mathbb{R}^{TH'W'\times TH'W'}$ indicates the key--query matching scores. This allows us to diminish the influence of the object regions even after the softmax operation. In summary, the MMHA plays two roles: 1) it prevents object regions from bringing semantic cues from other object regions; 2) it forces object regions not to affect other valid regions by limiting content propagation.

\vspace{1mm}
\noindent\textbf{Spatio-temporal transformer.} As shown in Figure~\ref{figure4}, we divide the general transformer block into a temporal transformer block (TTB) and a spatial transformer block (STB). The architectures of both are identical to that of the general transformer architecture, i.e., the sequential connection of multi-head attention and multi-layer perceptron. However, the TTB exchanges semantic cues between multiple frames, while the STB internally exchanges semantic cues in a single frame. Furthermore, due to the difference in the attention layer, they have different objectives. The TTB is used to erase object cues and fill object regions by leveraging the cues from other frames, while the STB is used to internally fill those regions within a single frame.

Given that $F_l$ is $l$-th features where $l \in \{0, 1, ..., L-1\}$, and $L$ is the number of total transformer blocks, the transformer block formula can be defined as
\begin{eqnarray}
    &F'_l=\mathbf{MMHA}(\mathbf{LN}(F_l), \hat{M}) + F_l~,\\
	&\hat{F}_l = \mathbf{MLP}(\mathbf{LN}(F'_l)) + F'_l~,\\
    &F''_l=\mathbf{SMHA}(\mathbf{LN}(\hat{F}_l)) + \hat{F}_l~,\\
	&F_{l+1} = \mathbf{MLP}(\mathbf{LN}(F''_l)) + F''_l~,
\end{eqnarray}
where $\mathbf{MLP}$ and $\mathbf{LN}$ indicate multi-layer perceptron and layer normalization layer~\cite{layernorm}, respectively.

\begin{table*}[t]
	\centering
	\caption{Quantitative evaluation on the synthesized datasets.}
	\begin{tabular}{P{1.5cm}|P{2cm}|P{1.2cm}P{1.2cm}P{1.2cm}|P{1.2cm}P{1.2cm}P{1.2cm}}
	    \multirow{2}{*}{VOS} & \multirow{2}{*}{VI} & \multicolumn{3}{c|}{DAYT} & \multicolumn{3}{c}{BLDA}\\
	    & & PSNR $\uparrow$ & SSIM $\uparrow$ & LPIPS $\downarrow$ & PSNR $\uparrow$ & SSIM $\uparrow$ & LPIPS $\downarrow$\\
	    \midrule
		\multirow{4}{*}{FRTM}& CPNet & 27.038 & 0.9324 & 0.0946 & 29.998 & 0.9644 & 0.0451\\
		& STTN & 25.357 & 0.9118 & 0.1032 & 28.167 & 0.9388 & 0.0552\\
	    & FGVC & 27.082 & 0.9300 & 0.0922 & 30.114 & 0.9650 & 0.0409\\
	    & FuseFormer & 28.495 & 0.9343 & 0.0901 & 34.115 & 0.9661 & 0.0422\\
		\midrule
		\multirow{4}{*}{CFBI}& CPNet & 31.985 & 0.9591 & 0.0616 & 33.716 & 0.9728 & 0.0336\\
		& STTN & 30.388 & 0.9400 & 0.0618 & 31.139 & 0.9477 & 0.0414\\
	    & FGVC & 32.124 & 0.9541 & 0.0484 & 34.491 & 0.9749 & \underline{0.0240}\\
	    & FuseFormer &\underline{34.309} &\underline{0.9623} &\textbf{0.0472} & 38.427 & 0.9751 & 0.0282\\
		\midrule
		\multirow{4}{*}{BMVOS}& CPNet & 31.910 & 0.9587 & 0.0596 & 33.405 & 0.9723 & 0.0325 \\
		& STTN & 30.173 & 0.9389 & 0.0615 & 30.799 & 0.9464 & 0.0420\\
	    & FGVC & 31.941 & 0.9523 & 0.0487 & 33.719 & 0.9737 & 0.0248\\
	    & FuseFormer & 34.055 & 0.9607 & 0.0480 & 36.679 & 0.9742 & 0.0289\\
		\midrule
		\multirow{4}{*}{TBD}& CPNet & 32.479 & 0.9597 & 0.0585 & 35.202 & 0.9755 & 0.0274\\
		& STTN & 30.469 & 0.9405 & 0.0600 & 32.021 & 0.9502 & 0.0362\\
	    & FGVC & 32.105 & 0.9533 & 0.0490 & 35.247 & 0.9766 & \textbf{0.0202}\\
	    & FuseFormer & 34.296 & 0.9618 & \underline{0.0479} &\underline{38.785} & \underline{0.9771} & 0.0246\\
	    \midrule
		\multicolumn{2}{c|}{Ours} &\textbf{35.518} &\textbf{0.9650} &0.0515 &\textbf{38.959} &\textbf{0.9783} & 0.0268\\
	\end{tabular}
	\label{comparison}
\end{table*}

\subsection{Loss Function}
To learn the mask prediction module, we apply mask loss for network training. It is a standard cross-entropy loss, as described in Eqn.~\ref{mask loss}, where $p$ indicates a single pixel location in $M$.
\begin{eqnarray}
	&L_{mask} = - \sum_p \log P(\hat{M} = M)
	\label{mask loss}      
\end{eqnarray}
For the video completion module, we use $L_1$ loss by separately applying it to the object and valid regions. We weight the loss based on the number of pixels in the area as each video has different ratios of object and valid regions. The losses are defined as
\begin{equation}
	L_{object} = \frac{\|M \odot (\hat{O}-O)\|_1}{\|M \|_1}~,
\end{equation}
\begin{equation}
    L_{valid} = \frac{\|(1-M) \odot (\hat{O}-O)\|_1}{\|1-M \|_1}~,
\end{equation}
where $\odot$ indicates Hadamard product. To ensure high perceptual quality, we also employ adversarial loss. The discriminator $D$ takes $O$ and $\hat{O}$ as its input, and outputs 1 if input seems real and 0 if input seems fake. By contrary, the network is trained make the discriminator make a wrong decision by generating a real-like fake image. The loss function for the discriminator is formulated as 
\begin{eqnarray}
    &L_{dis}=\mathbb{E}_O[\log D(O)]+\mathbb{E}_{\hat{O}}[\log(1-D(\hat{O}))]~,
\end{eqnarray}
where the loss function for the network is formulated as
\begin{eqnarray}
    &L_{adv}=\mathbb{E}_{\hat{O}}[\log D(\hat{O})].
\end{eqnarray}
In conclusion, total loss for our network can be defined as follows.
\begin{eqnarray}
    &L_{total} = L_{mask} + L_{object} + L_{valid} + L_{adv}
\end{eqnarray}

\subsection{Network Training}
We adopt two datasets for network training, COCO~\cite{COCO} and YouTube-VOS 2018~\cite{YTVOS}. As COCO is an image dataset, we randomly augment each image to generate videos, following the protocol in STM~\cite{STM} and TBD~\cite{TBD}. We resize all training videos to a $240 \times 432$ resolution and use them as clean videos. To generate input videos, we attach objects scrapped from other training snippets. The batch size is set to four and each snippet contains seven frames. We use an Adam optimizer~\cite{adam} for network optimization.

\section{Experiments}
\subsection{Experimental Setup}
\noindent\textbf{Datasets.} To validate the effectiveness of our method, we synthesize the testing videos using DAVIS~\cite{DAVIS2016, DAVIS2017}, YouTube-VOS 2018~\cite{YTVOS}, and BL30K~\cite{MiVOS} datasets. We design two synthesized datasets, DAYT and BLDA, which mean DAVIS objects on YouTube-VOS 2018 videos and BL30K objects on DAVIS videos. DAYT and BLDA consist of 88 and 120 video sequences, each consisting of input images, object masks, and output clean images.

\vspace{1mm}
\noindent\textbf{Detailed settings.} To compare our method with existing two-stage methods, we use state-of-the-art VI and VOS methods with publicly available pre-trained models. As DAVIS and BL30K are used to construct objects in our testing datasets, we do not adopt VOS models trained on those datasets. For a fair comparison, we use the official code and pre-trained models for all methods. All experiments are implemented on a single RTX A6000 GPU.

\begin{figure*}[t]
\centering
	\includegraphics[width=1\linewidth]{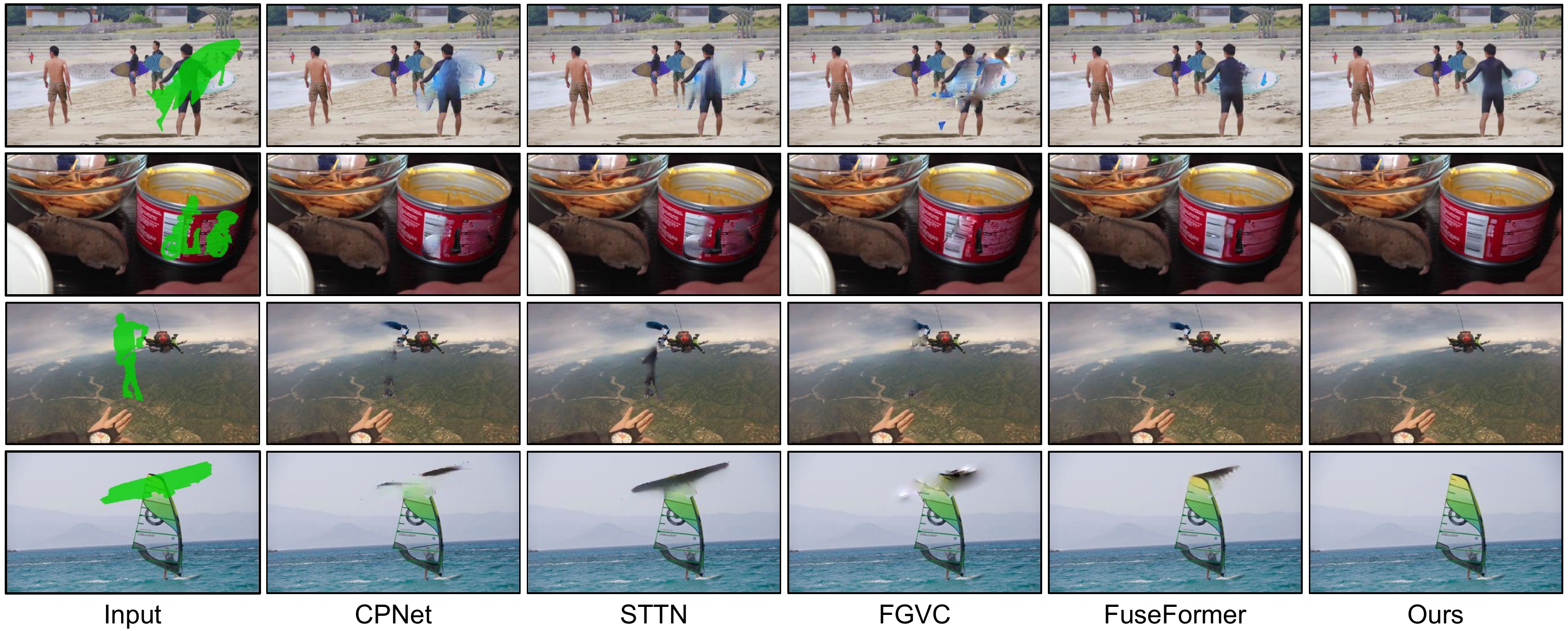}
    \caption{Qualitative comparison between state-of-the-art two-stage methods and our proposed method. Green regions indicate the object regions to be removed.}
    \label{figure5}
\end{figure*}

\begin{table*}[t]
    \centering
    \caption{Ablation study on our proposed algorithm on the synthesized datasets.}
    \begin{tabular}{P{3.4cm}|P{1.2cm}P{1.2cm}P{1.2cm}|P{1.2cm}P{1.2cm}P{1.2cm}}
         \multirow{2}{*}{Version} & \multicolumn{3}{c|}{DAYT} & \multicolumn{3}{c}{BLDA}\\
         & PSNR $\uparrow$ & SSIM $\uparrow$ & LPIPS $\downarrow$ & PSNR $\uparrow$ & SSIM $\uparrow$ & LPIPS $\downarrow$\\
         \midrule
         w/o mask loss &24.624 &0.9247 &0.1075 &28.218 &0.9561 &0.0589\\
         w/o mask guidance &33.298 &0.9558 &0.0638 &36.713 &0.9719 &0.0353\\
         SMHA $\rightarrow$ MMHA &26.970 &0.9287 &0.0955 &29.647 &0.9580 &0.0529\\
         no end-to-end training & 34.871 & 0.9633 & 0.0539 & 37.292 & 0.9762 & 0.0304 \\
         w/o encoder sharing & 32.324 & 0.9526 & 0.0694 & 35.790 & 0.9703 & 0.0375\\
         \midrule
         Ours &35.518 &0.9650 &0.0515 &38.959 &0.9783 &0.0268
    \end{tabular}
    \label{ablation}
\end{table*}

\subsection{Quantitative Results}
We quantitatively compare our method to existing two-stage methods in Table~\ref{comparison}. For the VOS models, we adopt FRTM~\cite{FRTM}, CFBI~\cite{CFBI}, BMVOS~\cite{BMVOS}, and TBD~\cite{TBD}. For the VI models, CPNet~\cite{CPNet}, STTN~\cite{STTN}, FGVC~\cite{FGVC}, and FuseFormer~\cite{liu2021fuseformer} are used. Among the VI models, FuseFormer shows the highest performance. When compared to other two-stage methods, it obtains the best accuracy on DAYT with CFBI, and on BLDA with TBD. We can also observe that if segmentation masks are not accurate enough, satisfactory OSVI performance cannot be achieved even if the state-of-the-art VI method is adopted. This supports the claim that a unified OSVI pipeline capable of effectively handling such noises or errors is needed. On both datasets, our method significantly outperforms all previous methods in PSNR and SSIM metrics. Quantitative results demonstrate the superiority of our method compared to state-of-the-art two-stage methods.

\subsection{Qualitative Results}
\begin{figure*}[h]
\centering
	\includegraphics[width=1\linewidth]{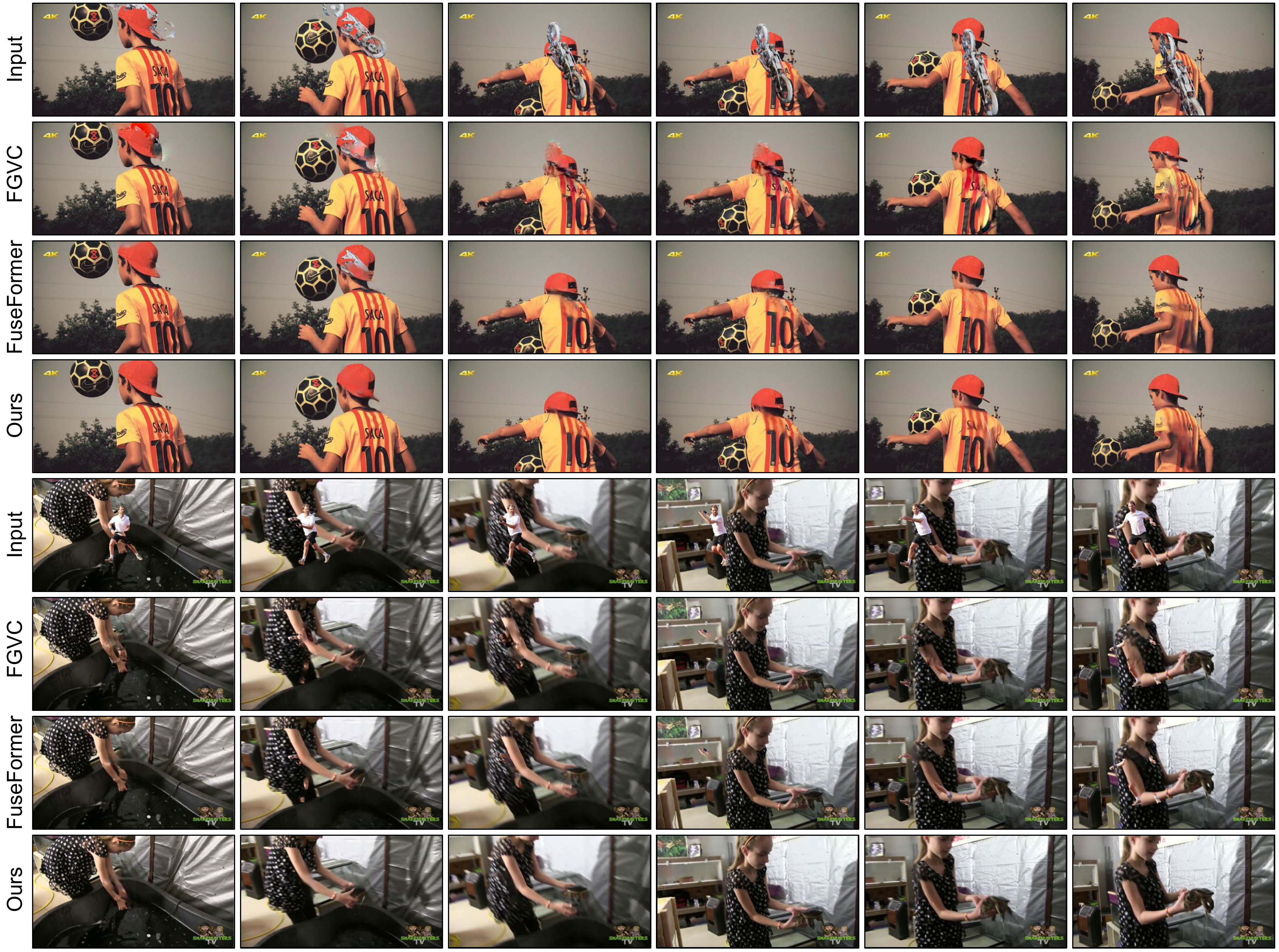}
    \caption{Example input and output videos of different one-shot video inpainting algorithms.}
    \label{figure6}
\end{figure*}

In Figure~\ref{figure5} and Figure~\ref{figure6}, we qualitatively compare our method to state-of-the-art two-stage methods. As a VOS model, TBD~\cite{TBD} is adopted for all VI models because it quantitatively outperforms other methods in Table~\ref{comparison}. From Figure~\ref{figure5}, we can conclude that optical flow-based methods, such as FGVC~\cite{FGVC}, are inadequate for OSVI. This is because un-erased objects may be copied and pasted to fill the object regions, and make the flow-based method produce the artifacts or some afterimages. Our method shows the clearest and most accurate object removal quality. Qualitative results are also compared in video form in Figure~\ref{figure6}.

\subsection{Ablation Study}
In Table~\ref{ablation}, we conduct an ablation study on various components of our method. Each different version is quantitatively compared on two synthesized datasets, DAYT and BLDA.

\vspace{1mm}
\noindent\textbf{Mask loss.} We conduct a study on the use of mask loss to figure out how mask supervision affects object removal quality. Removing mask loss leads to a drastic performance degradation as small errors in object regions lead to critical errors in the entire system. This demonstrates the need for mask supervision via mask loss is an essential component.

\vspace{1mm}
\noindent\textbf{Mask guidance}. To filter out object cues extracted from the feature encoder, we apply the mask guidance matrix in MMHA. The quantitative results demonstrate the effectiveness of this filtering process. Compared to the model version without filtering, the model version with filtering shows better performance on all metrics.

\vspace{1mm}
\noindent\textbf{Multi-head attention}. As TTB focuses on filling the object regions missing the object cues, we utilize STB to focus on filling the object regions that cannot be filled in TTB. Therefore, we apply the MMHA method only to TTB and not to STB. To prove its functionality, we evaluate the model version designed with MMHA instead of SMHA. It turns out that the employed masked attention was ineffective, in which it could not fill the object regions. Thus, the collaborative employment of masked attention for temporal information and naive attention for spatial information is required for OSVI.

\vspace{1mm}
\noindent\textbf{End-to-end training.} To demonstrate the effectiveness of end-to-end network training, we compare the models with and without the end-to-end training protocol. For the model without end-to-end training, the video completion module takes ground truth masks and extracted features without gradients to be separately trained from the mask prediction module. As observed from the table, end-to-end network training enables the network to be more effective. Meaningful improvements are observable for all metrics.

\vspace{1mm}
\noindent\textbf{Encoder sharing.} We also conduct an experiment to see the difference between models with single and multiple encoders. It is interesting to observe that using a single encoder shows much better performance than using separate encoders. This demonstrates that mask prediction and video completion share common complementary properties. Moreover, jointly learning them ensures a more appropriate network for OSVI.

\section{Conclusion}
Inspired by inefficiency of conventional VI methods for real-world applications, we deal with the task of OSVI only requiring a single-frame manual annotation. To better deal with this without merely sequentially connecting existing VOS and VI algorithms, we propose an end-to-end learnable baseline model. By a significant margin, our method outperforms all existing two-stage methods both quantitatively and qualitatively on two synthesized datasets. We believe our research makes a step towards efficient and applicable VI.

% Use \bibliography{yourbibfile} instead or the References section will not appear in your paper
\bibliography{aaai23}

\begin{thebibliography}{37}
\providecommand{\natexlab}[1]{#1}

\bibitem[{Ba, Kiros, and Hinton(2016)}]{layernorm}
Ba, J.~L.; Kiros, J.~R.; and Hinton, G.~E. 2016.
\newblock Layer normalization.
\newblock \emph{arXiv preprint arXiv:1607.06450}.

\bibitem[{Be{\v{s}}i{\'c} and Valada(2020)}]{DynaFill}
Be{\v{s}}i{\'c}, B.; and Valada, A. 2020.
\newblock Dynamic Object Removal and Spatio-Temporal RGB-D Inpainting via
  Geometry-Aware Adversarial Learning.
\newblock \emph{arXiv preprint arXiv:2008.05058}.

\bibitem[{Caelles et~al.(2017)Caelles, Maninis, Pont-Tuset, Leal-Taix{\'e},
  Cremers, and Van~Gool}]{OSVOS}
Caelles, S.; Maninis, K.-K.; Pont-Tuset, J.; Leal-Taix{\'e}, L.; Cremers, D.;
  and Van~Gool, L. 2017.
\newblock One-shot video object segmentation.
\newblock In \emph{Proceedings of the IEEE conference on computer vision and
  pattern recognition}, 221--230.

\bibitem[{Chang, Yu~Liu, and Hsu(2019)}]{VORNet}
Chang, Y.-L.; Yu~Liu, Z.; and Hsu, W. 2019.
\newblock Vornet: Spatio-temporally consistent video inpainting for object
  removal.
\newblock In \emph{Proceedings of the IEEE/CVF Conference on Computer Vision
  and Pattern Recognition Workshops}, 0--0.

\bibitem[{Cheng and Schwing(2022)}]{XMem}
Cheng, H.~K.; and Schwing, A.~G. 2022.
\newblock XMem: Long-Term Video Object Segmentation with an Atkinson-Shiffrin
  Memory Model.
\newblock \emph{arXiv preprint arXiv:2207.07115}.

\bibitem[{Cheng, Tai, and Tang(2021)}]{MiVOS}
Cheng, H.~K.; Tai, Y.-W.; and Tang, C.-K. 2021.
\newblock Modular Interactive Video Object Segmentation: Interaction-to-Mask,
  Propagation and Difference-Aware Fusion.
\newblock In \emph{CVPR}.

\bibitem[{Cho et~al.(2022{\natexlab{a}})Cho, Lee, Kim, Jang, and Lee}]{BMVOS}
Cho, S.; Lee, H.; Kim, M.; Jang, S.; and Lee, S. 2022{\natexlab{a}}.
\newblock Pixel-Level Bijective Matching for Video Object Segmentation.
\newblock In \emph{Proceedings of the IEEE/CVF Winter Conference on
  Applications of Computer Vision}, 129--138.

\bibitem[{Cho et~al.(2022{\natexlab{b}})Cho, Lee, Lee, Park, Jang, Kim, and
  Lee}]{TBD}
Cho, S.; Lee, H.; Lee, M.; Park, C.; Jang, S.; Kim, M.; and Lee, S.
  2022{\natexlab{b}}.
\newblock Tackling Background Distraction in Video Object Segmentation.
\newblock \emph{arXiv preprint arXiv:2207.06953}.

\bibitem[{Gao et~al.(2020)Gao, Saraf, Huang, and Kopf}]{FGVC}
Gao, C.; Saraf, A.; Huang, J.-B.; and Kopf, J. 2020.
\newblock Flow-edge guided video completion.
\newblock In \emph{European Conference on Computer Vision}, 713--729. Springer.

\bibitem[{Hu, Huang, and Schwing(2018)}]{VideoMatch}
Hu, Y.-T.; Huang, J.-B.; and Schwing, A.~G. 2018.
\newblock Videomatch: Matching based video object segmentation.
\newblock In \emph{Proceedings of the European Conference on Computer Vision
  (ECCV)}, 54--70.

\bibitem[{Huang et~al.(2016)Huang, Kang, Ahuja, and Kopf}]{huang}
Huang, J.-B.; Kang, S.~B.; Ahuja, N.; and Kopf, J. 2016.
\newblock Temporally coherent completion of dynamic video.
\newblock \emph{ACM Transactions on Graphics (TOG)}, 35(6): 1--11.

\bibitem[{Kim et~al.(2019)Kim, Woo, Lee, and Kweon}]{VINet}
Kim, D.; Woo, S.; Lee, J.-Y.; and Kweon, I.~S. 2019.
\newblock Deep video inpainting.
\newblock In \emph{Proceedings of the IEEE/CVF Conference on Computer Vision
  and Pattern Recognition}, 5792--5801.

\bibitem[{Kingma and Ba(2014)}]{adam}
Kingma, D.~P.; and Ba, J. 2014.
\newblock Adam: A method for stochastic optimization.
\newblock \emph{arXiv preprint arXiv:1412.6980}.

\bibitem[{Lee et~al.(2019)Lee, Oh, Won, and Kim}]{CPNet}
Lee, S.; Oh, S.~W.; Won, D.; and Kim, S.~J. 2019.
\newblock Copy-and-paste networks for deep video inpainting.
\newblock In \emph{Proceedings of the IEEE/CVF International Conference on
  Computer Vision}, 4413--4421.

\bibitem[{Li et~al.(2020)Li, Zhao, Ma, Gong, Qi, Zhang, Tao, and
  Kotagiri}]{short}
Li, A.; Zhao, S.; Ma, X.; Gong, M.; Qi, J.; Zhang, R.; Tao, D.; and Kotagiri,
  R. 2020.
\newblock Short-Term and Long-Term Context Aggregation Network for Video
  Inpainting.
\newblock In \emph{European Conference on Computer Vision}, 728--743. Springer.

\bibitem[{Li et~al.(2022)Li, Lu, Qin, Guo, and Cheng}]{li2022towards}
Li, Z.; Lu, C.-Z.; Qin, J.; Guo, C.-L.; and Cheng, M.-M. 2022.
\newblock Towards an end-to-end framework for flow-guided video inpainting.
\newblock In \emph{Proceedings of the IEEE/CVF Conference on Computer Vision
  and Pattern Recognition}, 17562--17571.

\bibitem[{Liang et~al.(2020)Liang, Li, Jafari, and Chen}]{AFB-URR}
Liang, Y.; Li, X.; Jafari, N.; and Chen, J. 2020.
\newblock Video object segmentation with adaptive feature bank and
  uncertain-region refinement.
\newblock \emph{Advances in Neural Information Processing Systems}, 33:
  3430--3441.

\bibitem[{Lin et~al.(2014)Lin, Maire, Belongie, Hays, Perona, Ramanan,
  Doll{\'a}r, and Zitnick}]{COCO}
Lin, T.-Y.; Maire, M.; Belongie, S.; Hays, J.; Perona, P.; Ramanan, D.;
  Doll{\'a}r, P.; and Zitnick, C.~L. 2014.
\newblock Microsoft coco: Common objects in context.
\newblock In \emph{European conference on computer vision}, 740--755. Springer.

\bibitem[{Liu et~al.(2021)Liu, Deng, Huang, Shi, Lu, Sun, Wang, Dai, and
  Li}]{liu2021fuseformer}
Liu, R.; Deng, H.; Huang, Y.; Shi, X.; Lu, L.; Sun, W.; Wang, X.; Dai, J.; and
  Li, H. 2021.
\newblock Fuseformer: Fusing fine-grained information in transformers for video
  inpainting.
\newblock In \emph{Proceedings of the IEEE/CVF International Conference on
  Computer Vision}, 14040--14049.

\bibitem[{Maninis et~al.(2018)Maninis, Caelles, Chen, Pont-Tuset,
  Leal-Taix{\'e}, Cremers, and Van~Gool}]{OSVOS-S}
Maninis, K.-K.; Caelles, S.; Chen, Y.; Pont-Tuset, J.; Leal-Taix{\'e}, L.;
  Cremers, D.; and Van~Gool, L. 2018.
\newblock Video object segmentation without temporal information.
\newblock \emph{IEEE transactions on pattern analysis and machine
  intelligence}, 41(6): 1515--1530.

\bibitem[{Oh et~al.(2019{\natexlab{a}})Oh, Lee, Xu, and Kim}]{STM}
Oh, S.~W.; Lee, J.-Y.; Xu, N.; and Kim, S.~J. 2019{\natexlab{a}}.
\newblock Video object segmentation using space-time memory networks.
\newblock In \emph{Proceedings of the IEEE International Conference on Computer
  Vision}, 9226--9235.

\bibitem[{Oh et~al.(2019{\natexlab{b}})Oh, Lee, Lee, and Kim}]{OPN}
Oh, S.~W.; Lee, S.; Lee, J.-Y.; and Kim, S.~J. 2019{\natexlab{b}}.
\newblock Onion-peel networks for deep video completion.
\newblock In \emph{Proceedings of the IEEE/CVF International Conference on
  Computer Vision}, 4403--4412.

\bibitem[{Perazzi et~al.(2016)Perazzi, Pont-Tuset, McWilliams, {Van Gool},
  Gross, and Sorkine-Hornung}]{DAVIS2016}
Perazzi, F.; Pont-Tuset, J.; McWilliams, B.; {Van Gool}, L.; Gross, M.; and
  Sorkine-Hornung, A. 2016.
\newblock A Benchmark Dataset and Evaluation Methodology for Video Object
  Segmentation.
\newblock In \emph{Computer Vision and Pattern Recognition}.

\bibitem[{Pont-Tuset et~al.(2017)Pont-Tuset, Perazzi, Caelles, Arbel\'aez,
  Sorkine-Hornung, and {Van Gool}}]{DAVIS2017}
Pont-Tuset, J.; Perazzi, F.; Caelles, S.; Arbel\'aez, P.; Sorkine-Hornung, A.;
  and {Van Gool}, L. 2017.
\newblock The 2017 DAVIS Challenge on Video Object Segmentation.
\newblock \emph{arXiv:1704.00675}.

\bibitem[{Robinson et~al.(2020)Robinson, Lawin, Danelljan, Khan, and
  Felsberg}]{FRTM}
Robinson, A.; Lawin, F.~J.; Danelljan, M.; Khan, F.~S.; and Felsberg, M. 2020.
\newblock Learning Fast and Robust Target Models for Video Object Segmentation.
\newblock In \emph{IEEE/CVF Conference on Computer Vision and Pattern
  Recognition (CVPR)}.

\bibitem[{Shetty, Fritz, and Schiele(2018)}]{Auto2018adversarial}
Shetty, R.; Fritz, M.; and Schiele, B. 2018.
\newblock Adversarial scene editing: Automatic object removal from weak
  supervision.
\newblock \emph{arXiv preprint arXiv:1806.01911}.

\bibitem[{Vaswani et~al.(2017)Vaswani, Shazeer, Parmar, Uszkoreit, Jones,
  Gomez, Kaiser, and Polosukhin}]{attention}
Vaswani, A.; Shazeer, N.; Parmar, N.; Uszkoreit, J.; Jones, L.; Gomez, A.~N.;
  Kaiser, {\L}.; and Polosukhin, I. 2017.
\newblock Attention is all you need.
\newblock \emph{Advances in neural information processing systems}, 30.

\bibitem[{Voigtlaender et~al.(2019)Voigtlaender, Chai, Schroff, Adam, Leibe,
  and Chen}]{FEELVOS}
Voigtlaender, P.; Chai, Y.; Schroff, F.; Adam, H.; Leibe, B.; and Chen, L.-C.
  2019.
\newblock Feelvos: Fast end-to-end embedding learning for video object
  segmentation.
\newblock In \emph{Proceedings of the IEEE Conference on Computer Vision and
  Pattern Recognition}, 9481--9490.

\bibitem[{Voigtlaender and Leibe(2017)}]{OnAVOS}
Voigtlaender, P.; and Leibe, B. 2017.
\newblock Online adaptation of convolutional neural networks for the 2017 davis
  challenge on video object segmentation.
\newblock In \emph{The 2017 DAVIS Challenge on Video Object Segmentation-CVPR
  Workshops}, volume~5.

\bibitem[{Wexler, Shechtman, and Irani(2004)}]{space}
Wexler, Y.; Shechtman, E.; and Irani, M. 2004.
\newblock Space-time video completion.
\newblock In \emph{Proceedings of the 2004 IEEE Computer Society Conference on
  Computer Vision and Pattern Recognition, 2004. CVPR 2004.}, volume~1, I--I.
  IEEE.

\bibitem[{Woo et~al.(2018)Woo, Park, Lee, and Kweon}]{CBAM}
Woo, S.; Park, J.; Lee, J.-Y.; and Kweon, I.~S. 2018.
\newblock Cbam: Convolutional block attention module.
\newblock In \emph{Proceedings of the European conference on computer vision
  (ECCV)}, 3--19.

\bibitem[{Xu et~al.(2018)Xu, Yang, Fan, Yue, Liang, Yang, and Huang}]{YTVOS}
Xu, N.; Yang, L.; Fan, Y.; Yue, D.; Liang, Y.; Yang, J.; and Huang, T. 2018.
\newblock Youtube-vos: A large-scale video object segmentation benchmark.
\newblock \emph{arXiv preprint arXiv:1809.03327}.

\bibitem[{Xu et~al.(2019)Xu, Li, Zhou, and Loy}]{DFVI}
Xu, R.; Li, X.; Zhou, B.; and Loy, C.~C. 2019.
\newblock Deep flow-guided video inpainting.
\newblock In \emph{Proceedings of the IEEE/CVF Conference on Computer Vision
  and Pattern Recognition}, 3723--3732.

\bibitem[{Yang, Wei, and Yang(2020)}]{CFBI}
Yang, Z.; Wei, Y.; and Yang, Y. 2020.
\newblock Collaborative video object segmentation by foreground-background
  integration.
\newblock \emph{arXiv preprint arXiv:2003.08333}.

\bibitem[{Zeng, Fu, and Chao(2020)}]{STTN}
Zeng, Y.; Fu, J.; and Chao, H. 2020.
\newblock Learning Joint Spatial-Temporal Transformations for Video Inpainting.
\newblock In \emph{European Conference on Computer Vision}, 528--543. Springer.

\bibitem[{Zhang et~al.(2020)Zhang, Li, Wang, Guan, Fang, Song, Yu, Chen, Xu,
  and Yang}]{autoremover}
Zhang, R.; Li, W.; Wang, P.; Guan, C.; Fang, J.; Song, Y.; Yu, J.; Chen, B.;
  Xu, W.; and Yang, R. 2020.
\newblock Autoremover: automatic object removal for autonomous driving videos.
\newblock In \emph{Proceedings of the AAAI Conference on Artificial
  Intelligence}, volume~34, 12853--12861.

\bibitem[{Zou et~al.(2021)Zou, Yang, Liu, and Lee}]{TSAM}
Zou, X.; Yang, L.; Liu, D.; and Lee, Y.~J. 2021.
\newblock Progressive temporal feature alignment network for video inpainting.
\newblock In \emph{Proceedings of the IEEE/CVF Conference on Computer Vision
  and Pattern Recognition}, 16448--16457.

\end{thebibliography}

\end{document}

\maketitle

\begin{table*}[t]
    \centering
    \caption{Quantitative mask quality comparison of various models.}
    \begin{tabular}{P{5cm}|P{2cm}P{2cm}|P{2cm}P{2cm}}
         \multirow{2}{*}{Version} & \multicolumn{2}{c|}{DAYT} & \multicolumn{2}{c}{BLDA}\\
         &IoU $\uparrow$ &Recall $\uparrow$ &IoU $\uparrow$ &Recall $\uparrow$\\
         \midrule
         Ours &81.6 &91.1 &80.4 &89.2\\
        %  \midrule
        %  Ours (no end-to-end training) &78.4 (-3.2) &88.3 (-2.8) &72.7 (-7.7) &80.3 (-8.9)\\
        %  Ours (w/o encoder sharing) &77.5 (-4.1) &85.6 (-5.5) &74.6 (-5.8) &81.9 (-7.3)\\
         \midrule
         BMVOS~\cite{BMVOS} &75.4 (-6.2) &84.2 (-6.9) &68.1 (-12.3) &77.0 (-12.2)\\
         CFBI~\cite{CFBI} &77.3 (-4.3) &82.4 (-8.7) &71.0 (-9.4) &76.2 (-13.0)\\
         TBD~\cite{TBD} &79.5 (-2.1) &86.0 (-5.1) &77.8 (-2.6) &84.9 (-4.3)\\
    \end{tabular}
    \label{comp}
\end{table*}

\section{Additional Study on Mask Prediction}
As OSVI is a task to remove a designated object completely and fill in the missing regions, mask prediction is an important process for accurate OSVI. Here, we conduct an additional study on the mask prediction quality.

\subsection{Quantitative Study}
We evaluate the mask prediction quality of our method quantitatively. As the evaluation metrics, we adopt IoU and Recall. IoU is a general metric for evaluating segmentation performance, and can be obtained by the area of overlap over the area of union. Recall is a metric that calculates how much the ground truth region is sufficiently detected, and can be obtained by the area of overlap over the area of the ground truth. Considering OSVI is a task to completely remove a designated object, it will bring a severe error to the system if a tracker fails to track the object even in part. Therefore, unlike general segmentation tasks, Recall metric should be considered the major metric in OSVI.

In Table~\ref{comp}, our method is compared to state-of-the-art VOS methods quantitatively on the DAYT and BLDA datasets. Following previous settings, we adopt BMVOS~\cite{BMVOS}, CFBI~\cite{CFBI}, and TBD~\cite{TBD} as the compared VOS solutions. Our method outperforms all other methods by a significantly margin. This is because of a joint learning of mask prediction and video completion modules, which can help each others' robust representation learning. It is also interesting that compared to other VOS methods, the predicted masks of our method have higher Recall scores compared to IoU scores. As the mask prediction module is trained considering the video completion module, it tends to predict the masks in a safe way (large masks), rather than a strict way (accurate masks).

\subsection{Qualitative Study}
\begin{figure*}[t]
\centering
	\includegraphics[width=1\linewidth]{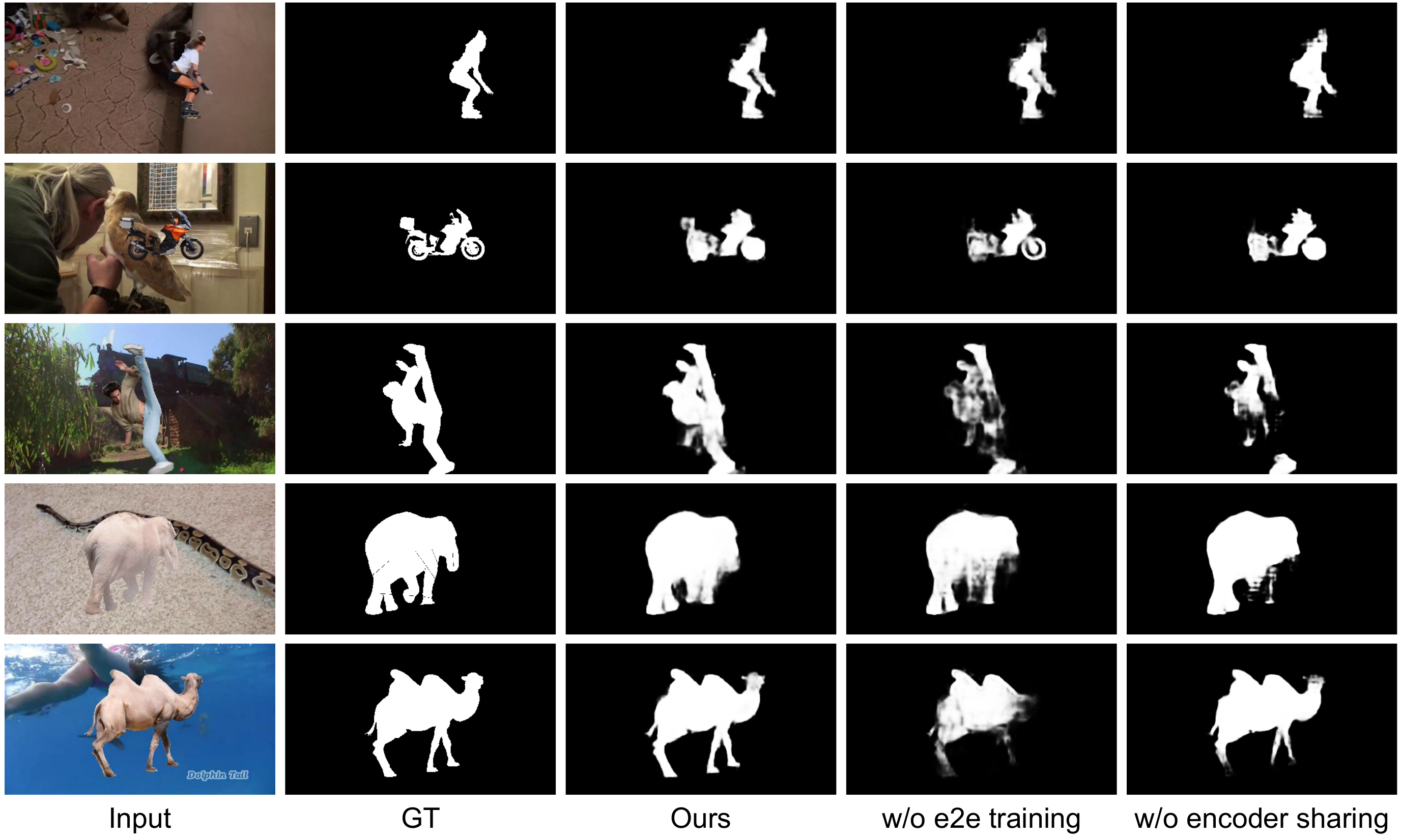}
    \caption{Qualitative mask comparison between various model versions.}
    \label{figure7}
\end{figure*}

In Figure~\ref{figure7}, we also visualize the qualitative comparison of different model versions of our method. We compare the model without end-to-end training scheme (intentional disconnection of gradients between mask prediction module and video completion module), the model without encoder sharing (separate encoders for mask prediction and video completion), and the final model.

Compared the the model without end-to-end training, our final model outputs more accurate and stable masks. As described before, the mask prediction module tends to generate large masks instead of fine and accurate masks. This is because even a small lost of an object can bring a severe breakdown of a system. Our final model also shows better segmentation performance compared to the model that employs separate encoders for mask prediction module and video completion module. This demonstrates the effectiveness of a joint learning of mask prediction and video completion modules. As mask prediction and video completion share common complementary properties, sharing a single encoder enables a stable network training and strong feature representation learning.

\bibliography{aaai23}